\documentclass[twoside]{article}
\pdfoutput=1 
\usepackage{aistats2019}

\usepackage[utf8]{inputenc}
\usepackage[english]{babel}
\usepackage{amsmath}
\usepackage[retainorgcmds]{IEEEtrantools}
\usepackage{mathrsfs}
\usepackage{graphicx} 
\usepackage{amssymb}
\usepackage{bm}
\usepackage{dsfont}
\graphicspath{{figure/}}

\usepackage{color,soul}
\usepackage{textcomp}

\usepackage{hyperref}
\usepackage[round]{natbib}
\begin{document}
	
	%
	
	%
	
	\twocolumn[
	\aistatstitle{Fast Exact Computation of Expected HyperVolume Improvement}
	
	
	\aistatsauthor{ Guang Zhao$^1$ \And Raymundo Arróyave$^2$ \And  Xiaoning Qian$^1$ }
	\aistatsaddress{$^1$Department of Electrical and Computer Engineering, Texas A\&M University\\ $^2$Department of Materials Science and Engineering, Texas A\&M University
	} 
	
	]
	\title{Fast Exact Computation of Expected HyperVolume Improvement}


	\begin{abstract}
		In multi-objective Bayesian optimization and surrogate-based evolutionary algorithms, Expected HyperVolume Improvement~(EHVI) is widely used as the acquisition function to guide the search approaching the Pareto front. This paper focuses on the exact calculation of EHVI given a nondominated set, for which the existing exact algorithms are complex and can be inefficient for problems with more than three objectives. Integrating with different decomposition algorithms, we propose a new method for calculating the integral in each decomposed high-dimensional box in constant time. We develop three new exact EHVI calculation algorithms based on three region decomposition methods. The first grid-based algorithm has a complexity of  $O(m\cdot n^m)$ with $n$ denoting the size of the nondominated set and $m$ the number of objectives. The Walking Fish Group (WFG)-based algorithm has a worst-case complexity of $O(m\cdot 2^n)$ but has a better average performance. These two can be applied for problems with any $m$. The third CLM-based \citep{beume2009complexity} algorithm is only for $m=3$ and asymptotically optimal with complexity $\Theta(n\log{n})$. Performance comparison results show that all our three algorithms are at least twice faster than the state-of-the-art algorithms with the same decomposition methods. When $m>3$, our WFG-based algorithm can be over $10^2$ faster than the corresponding existing algorithms.  Our algorithm is demonstrated in an example involving efficient multi-objective material design with Bayesian optimization. 
	\end{abstract}
	
	\section{Introduction}
	
	With rapid advancements in sensing and computing, artificial intelligence~(AI), especially machine learning~(ML), has become an important tool in a wide variety of applications. With available ``big'' data, AI/ML can help derive accurate prediction for better decision making to maximize the gain with reduced time and cost~\citep{mcafee2012big, provost2013data}. However, in applications involving complex systems, for example, in multi-functional materials discovery~\citep{cadman2013design}, it may not be possible to always have big data for effective ML. Often, model and data uncertainty, with limited training samples, have to be incorporated in a ``smart'' way for optimization to actively learn about the system and achieve the desirable goals. In such ``data-poor'' environments, Bayesian experimental design~\citep{chaloner1995bayesian} and reinforcement learning~\citep{wiering2012reinforcement} are adopted. 
	
	When simulations and experiments of a complex system under study are resource (cost and time) demanding, surrogate-based methods are adopted considering uncertainty for Bayesian experimental design. For example, Gaussian process models have been studied extensively as the surrogate models for the objective functions for decision making~\citep{frazier2009knowledge}. Based on that, the expected utility can be computed based on the application-specific acquisition function for Bayesian optimization to guide the search for decision making towards promising or less explored regions of the search space. We focus on Bayesian experimental design with multiple objective functions, for which Expected HyperVolume Improvement (EHVI) \citep{emmerich2006single} has proved to be an effective acquisition function~\citep{couckuyt2014fast}. Moreover, EHVI is also used for preselection criteria for multi-objective evolutionary algorithms \citep{luo2015study}. 
	
	
	
	Computing EHVI can be computationally demanding, especially when the number of objective functions and the complexity of the Pareto front increases. This is not desirable in many Bayesian experimental design problems. For example, for novel materials discovery, one of the major goals of applying Bayesian optimization is to reduce the time to run high-throughput simulations to identify candidate desirable materials. We need to guarantee that the computational time and complexity of Bayesian optimization is significantly better than those of high-throughput simulations to achieve our goals. Hence, efficiently computing EHVI for multi-objective Bayesian optimization is critical. 

	
	Usually EHVI is a multi-modal function in the search space. We need to search for the point with the maximum EHVI in each iteration of Bayesian optimization. As a result, EHVI needs to be calculated multiple times. Exact calculation for EHVI has been studied in the literature \citep{couckuyt2014fast, hupkens2015faster, emmerich2016multicriteria, yang2017computing}. Existing exact EHVI calculation algorithms have two steps: at the first step, the integral region is decomposed into high-dimensional boxes; and then at the second step the integral in each box is computed and added together to compute EHVI. The integral region decomposition methods have been well studied using the algorithms calculating hypervolume indicator~\citep{lacour2017box}. Existing EHVI calculation algorithms mostly tailor the integral computation in boxes to different decomposition methods and it is difficult to transfer across different decomposition methods~\citep{hupkens2015faster,yang2017computing}. For two- and three-objective cases, \cite{emmerich2016multicriteria, yang2017computing} have shown that EHVI can be calculated in asymptotically optimal time of $\Theta(n\log n)$ with $n$ denoting the size of the nondominated set. However, these algorithms cannot be extended for more than three objectives. The existing algorithms are complicated and time-consuming when $n$ is large with the number of objectives $m>3$ \citep{couckuyt2014fast}. 
	
	In this paper, we develop algorithms for the fast, exact computation of EHVI with different integral region decomposition methods. For any decomposition method given a nondominated set, we first transform the integral expression of EHVI and show that the transformed integrand has a closed form in each decomposed box and can be computed in constant time. We develop three different exact EHVI algorithms following this new scheme and name them after the decomposition methods. The first grid-based algorithm has a complexity of $O(m\cdot n^m)$, with $n$ denoting the size of the nondominated set and $m$ the number of objectives. The second WFG-based algorithm has a worst case complexity of $O(m\cdot 2^n)$ but experiments show it has a better average performance than grid-based algorithm. These two algorithms can be applied for problems with any $m$. A CLM-based algorithm is proposed merely for three objective problems with an asymptotically optimal complexity of $\Theta(n\log{n})$. 
	

	
	The paper is structured as follows: Section \ref{sec:EHVI} introduces the definition of EHVI and state-of-the-art exact calculation algorithms. Section \ref{sec:computation} derives another EHVI integral expression and the closed form of the integral in each box. Section \ref{sec:integral region} details three decomposition methods. Finally, Section \ref{sec:experiment} presents the performance comparison of the proposed algorithms with state-of-the-art exact algorithms and the proposed algorithm application on Bayesian optimization for multi-objective material design.

	\section{Background}\label{sec:EHVI}
	
	\subsection{Multi-objective Bayesian Optimization}
	Formulate $m$-objective minimization problem as: 
	\begin{equation}
	\min (f_1(\bm{x}), f_2(\bm{x}), \ldots, f_m(\bm{x})), \bm{x}\in \mathcal{X},
	\end{equation}
	where $\mathcal{X}$ is the feasible set or search space to explore. Denote the $m$-objective function image as $\mathcal{Y} \subset \mathbb{R}^m$. Usually there would not be a single solution optimizing every objective simultaneously. We aim to find the set of optimal trade-off solutions known as the Pareto front, which is defined as $\mathcal{Y}_p = \{\boldsymbol{y}\in\mathcal{Y}: \nexists \boldsymbol{y}' \, \text{s.t.} \, \boldsymbol{y}'\prec \boldsymbol{y} \}$. Here,  $\bm{y} \prec \bm{y}'$ reads as $\bm{y}$ dominates $\bm{y}'$ in the context of minimization, meaning that $\forall i\le m, y_i\le y'_i$ and $\exists j \le m, y_j < y'_j$. A vector $\bm{y}$ is nondominated with respect to a set $A$ if and only if $\bm{y}$ is not dominated by any vector in $A$. The nondominated set of $A$ is defined as the subset that includes all nondominated vectors with respect to $A$. So the Pareto front $\mathcal{Y}_p$ is the non-dominated set of $\mathcal{Y}$. 
	On the other hand, all the vectors not in the $\mathcal{Y}_p$ are dominated by at least one element in the $\mathcal{Y}_p$. So the Pareto front dominates the largest region of vectors in the objective space. In multi-objective optimization, we can use the dominated hypervolume as a metric guiding the search for the Pareto front as detailed below. 
	
	
	When the objective functions are unknown and expensive to evaluate, we wish to obtain an approximation to the Pareto front with the minimum number of required objective function evaluations. Bayesian Optimization (BO) is one solution to this problem. 
	BO is a sequential sampling method that samples new points utilizing the information of previously observed data. In each sampling iteration, a probabilistic model and an acquisition function are calculated. The probabilistic model describes the distribution of objective functions. It starts with a prior belief of objective functions and updates the belief based on the observed data, following Bayes' rule. Usually the surrogate models are chosen as independent Gaussian process models for their good interpolation performance and easy computation property~\citep{emmerich2008computation, picheny2015multiobjective}. 
	
	Given a Gaussian process prior, the posterior of the objective functions of a candidate point is still Gaussian. Hence, the computation of the acquisition function can be efficient.  The acquisition function describes the utility of querying each candidate point given the probabilistic model, usually considering both the utility of exploration and exploitation. The maximum point of the acquisition function is chosen to evaluate in each iteration. Compared with the original objective functions, the acquisition function is much cheaper to evaluate and maximize. In multi-objective problems, a popular choice of the acquisition function is EHVI for its good benchmark performance \citep{couckuyt2014fast}, which describes the expected gain in the dominated hypervolume. Its formal definition is given in the  following subsection. 
	
	\subsection{Expected HyperVolume Improvement}
	
	Assume $A\subset \mathcal{Y}$ is an objective vector set, the dominated hypervolume of  $A$ is defined as the hypervolume of the dominated region: 
	\begin{equation}
	\mathcal{H}(A) =\mathrm{Vol} \left( \{\bm{y}\in \mathbb{R}^m|\boldsymbol{y}\prec \boldsymbol{r}\, \text{and}\, \exists\, a\in A: \, a\prec \boldsymbol{y}\}\right),
	\label{eq:hypervolume}
	\end{equation}
	where $\boldsymbol{r}$ is a vector dominated by all the vectors in image $\mathcal{Y}$, called the reference point, which is introduced to bound the objective space, such that the dominated hypervolume is finite. We can also define the nondominated region as the whole region bounded by $\bm{r}$ minus the dominated region. From~\eqref{eq:hypervolume}, it's easy to see that the dominated hypervolume of $A$ equals to the dominated hypervolume of its nondominated set and therefore the dominated hypervolume of $\mathcal{Y}_p$ has the maximum dominated hypervolume, which equals to the dominated hypervolume of $\mathcal{Y}$. 
	Now for BO the problem of approaching the Pareto front can be transformed to finding the set with the maximum dominated hypervolume. 
	
	With the definition of dominated hypervolume, the hypervolume improvement of a vector $\bm{y}$ with respect to $A$ is defined as:	 
	\begin{equation}
	\mathcal{H}_I(\bm{y}, A) = \mathcal{H}(A\cup\{\bm{y}\}) - \mathcal{H}(A).
	\label{eq:HVI}
	\end{equation}
	Notice that the hypervolume improvement can only be nonzero when $\bm{y}$ falls in the nondominated region of $A$.  
	
	In the context of BO, $A$ is the observed set of the objective vectors and $\bm{y}$ is the random output of the probabilistic model corresponding to an candidate point. As a result, $\mathcal{H}_I(\bm{y}, A)$ is a random variable and EHVI is defined as its expected value over distribution  $P(\bm{y})$:
	\begin{equation}
	E\mathcal{H}_I(\bm{y}) = \mathbb{E}[\mathcal{H}_I(\bm{y}, A) ]=  \int_U \mathcal{H}_I( \bm{y}', A) P(\bm{y} = \bm{y}')\mathrm{d}\bm{y}', 	 					\label{eq:EHVI}
	\end{equation}
	which is an integral of hypervolume improvement over the nondominated region $U$ since in the dominated region $\mathcal{H}_I(\bm{y}', A) = 0$. EHVI is the expected gain in dominated hypervolume of observing one candidate point. Hence, in a myopic sequential sampling method, we will choose the candidate point with the maximum EHVI as the next point to query. 
	
	\subsection{Algorithms for Calculating EHVI}
	
	As shown in definition, EHVI is  the integral of hypervolume improvement with respect to the posterior measure over the nondominated region of a nondominated set. Since the nondominated region can have a complex shape, most existing exact EHVI calculation algorithms adopt a two-step approach, in which the first step is to decompose the integral region into high-dimensional boxes based on the observed set, and the second step is to calculate the integral contribution in each box separately and then add all the contributions together. For the integral in the second step, the integrand, i.e. the hypervolume improvement, also has a complex form that depends on the shape of the nondominated region. The integrand is often partitioned into terms by region decomposition so that the integral of each term has a closed form over each box for efficient computation. The complexity of such two-step approaches mainly depends on  $m$ and $n$ as larger $ m $ and $ n $ result in a more complex shape of nondominated regions.  
	
	
	\citet{couckuyt2014fast}  derived an algorithm (CDD13) capable of calculating the exact EHVI for any number of objectives. It obtains the nondominated region by subtracting the dominated region from the whole objective space and then recursively decomposes the dominated region into boxes following the Walking Fish Group (WFG) algorithm~\citep{while2012fast}, which calculates the exact hypervolumes. To calculate the integral in each box, CDD13 algorithm again decomposes the integrand, namely the hypervolume improvement into contributing terms belonging to every decomposed box, so that each term has a closed form integral over box. This nested decomposition scheme limits its application when $n$ and $m$ are large and the paper did not provide a complexity bound for the algorithm. 
	
	
	
	
	\citet{hupkens2015faster} proposed an algorithm (IRS) valid for calculating EHVI for three objectives in $O(n^3)$. IRS generates a grid structure based on the coordinates of the nondominated set, and decomposes the integral region into the corresponding $O(n^3)$ hyperrectangular grid cells~(boxes). Then the algorithm decomposes the integrand into 8 terms in 3d space. Thanks to the grid structure, the integral terms between some of different boxes share same multiplication factors, and the algorithm can calculate all the factors in advance within $O(n^3)$ time and store the calculation results in an array of size $O(n^2)$. However, since the integrand decomposition is complicated, the complexity of the algorithm has a large constant factor and it is hard to extend the algorithm for the problems with more than three objectives.
	
	More recently, \cite{yang2017computing}  developed a more efficient algorithm (KMAC) calculating EHVI for three objectives with the asymptotically optimal time complexity $\Theta(n\log n)$. The algorithm decomposes the integral region into $2n+1$ boxes, following an asymptotically optimal algorithm for calculating hypervolume improvement~\citep{emmerich2011computing}. With the integrand partitioned the same way as in \citet{couckuyt2014fast}, the algorithm takes advantage of the integral region decomposition method computing all the contributing hypervolumes in constant time. 
	However, this algorithm is only asymptotically optimal for computing a three-objective EHVI. Both the decomposition method and integral trick are difficult to be extended to problems with $m>3$. 
	
	Finally, \cite{feliot2017bayesian} wrote the formula for computing EHVI~(\ref{eq:EHVI}): 
	\begin{equation}
	E\mathcal{H}_I(\bm{y}) = \int_U P(\bm{y} \prec \bm{y}')\mathrm{d}\bm{y}',
	\label{eq:EHVI2}
	\end{equation}
	where $ P(\bm{y} \prec \bm{y}')$ is the probability that the random vector $\bm{y}$ is dominated by $\bm{y}'$. The integrand in this equation has a relatively simple form, which motivates our research. We will derive in the next section a closed-form expression. However, \citet{feliot2017bayesian} did not provide an exact algorithm calculating the integral but used a Monte Carlo sampling method instead. 
	
	In this paper, motivated by the above methods, we derive a closed-form integral computation for exact computation of~\eqref{eq:EHVI2}. With that, different decomposition methods can be integrated to derive more efficient exact EHVI calculating algorithms.

	\section{Closed-form Integral Computation} \label{sec:computation}

	In this section, we first show the equivalence of~\eqref{eq:EHVI} and~\eqref{eq:EHVI2}. In~\eqref{eq:HVI}, the hypervolume improvement is the hypervolume of the region that is dominated by $\bm{y}$ but not dominated by $A$.  Hence, it can be expressed with the indicator function $\mathds{1}(\cdot)$ as :
	\begin{equation}
	\mathcal{H}_I(\bm{y}, A) = \int_U \mathds{1}(\bm{y} \prec \bm{y}') \mathrm{d}\bm{y}'.
	\label{eq:HVI2}
	\end{equation}	
	Computing EHVI by~\eqref{eq:EHVI} can be transformed into another form \citep{feliot2017bayesian}:
	\begin{IEEEeqnarray}{rCl}
		E\mathcal{H}_I(\bm{y}) & = &  \mathbb{E}_{\bm{y}}\Big[  \int_U \mathds{1}(\bm{y} \prec \bm{y}') \mathrm{d}\bm{y}'\Big]\nonumber\\
		& = & \int_U \mathbb{E}_{\bm{y}}[\mathds{1}(\bm{y} \prec \bm{y}')]\mathrm{d}\bm{y}'\nonumber\\
		& = &\int_U P(\bm{y} \prec \bm{y}')\mathrm{d}\bm{y}',
		\label{eq:EHVIprob}
	\end{IEEEeqnarray}
	where in the second line we reverse the order of integrations based on Fubini's theorem, and the third line by the probability identity. This equation transfers the problem of computing EHVI into computing the integral of the cumulative probability $P(\bm{y} \prec \bm{y}')$ over the whole objective space, which is simpler as it does not depend on the shape of the nondominated region.
	
	With independent Gaussian process models as surrogate models for each objective in BO, the posterior output of each Gaussian process given the data can be considered as mutually independent Gaussian random variables $y_j \sim \mathcal{N}(\mu_j, \sigma_j^2), j\le m$, with $\mu_j, \sigma_j^2$ denoting the mean and covariance of j-th objective, respectively. Denote the probability distribution function of the standard normal distribution as $\phi(\cdot)$ and the corresponding cumulative distribution function as  $\Phi (\cdot)$. Given a new candidate point,  we have: 
	\begin{equation}
	P(\bm{y}\prec\bm{y}') = \prod_{j = 1}^{m} \Phi\Big(\frac{y'_j-\mu_j}{\sigma_j}\Big).
	\label{eq:8}
	\end{equation}	
	
	As we mentioned in Section~\ref{sec:EHVI}, the integral region $U$ can be decomposed into a set of high-dimensional boxes. Here we derive the closed-form integral computation of $ P(\bm{y}\prec\bm{y}')  $ in each box. Let $box(\bm{l}, \bm{u})$ indicate the half open region $(l_1, u_1]\times(l_2, u_2]\times\cdots\times(l_m, u_m]$, with $\bm{l} = (l_1, l_2, \ldots, l_m)$ denoting the lower bound and  $\bm{u} = (u_1, u_2, \ldots, u_m)$ denoting the upper bound. The integral of $P(\bm{y}\prec\bm{y}') $ in $box(\bm{l}, \bm{u})$ is:
	\begin{IEEEeqnarray}{rCl}
		\delta(box(\bm{u},\bm{l})) &=& \int_{box(\bm{u},\bm{l})}\prod_{j= 1}^{m}\Phi\Big(\frac{y_j'-\mu_j}{\sigma_j}\Big)\mathrm{d}\bm{y}'\nonumber\\
		& = & \prod_{j = 1}^{m}\int_{l_j}^{u_j}\Phi\Big(\frac{y_j'-\mu_j}{\sigma_j}\Big)\mathrm{d}y_j'.
		\label{eq:EHVIbox}
	\end{IEEEeqnarray}
	The integral of $\Phi(a+bx)$ has a closed form: $\int_{-\infty}^{t}\Phi(a+bx)\mathrm{d}x = \frac{1}{b}[(a+bt)\Phi(a+bt)+\phi(a+bt)]$ \citep{patel1996handbook}. 
	We then can write \eqref{eq:EHVIbox} as:
	\begin{IEEEeqnarray}{rCl}
		\delta(box(\bm{u},\bm{l})) & = & \prod_{j = 1}^{m}\left[\Psi(u_j, \mu_j, \sigma_j)-\Psi(l_j, \mu_j, \sigma_j)\right]
		\label{eq:boxclose}
	\end{IEEEeqnarray}
	with $\Psi(a, \mu, \sigma)  =  \int_{-\infty}^a\Phi(\frac{y-\mu}{\sigma})\mathrm{d}y = (a-\mu)\Phi(\frac{a-\mu}{\sigma})+\sigma\phi(\frac{a-\mu}{\sigma}).$
	
	The integral in one box in~\eqref{eq:boxclose} is not dependent on the other boxes or the nondominant set $A$. It can be calculated with the  computation complexity in $O(m)$, which is constant with respect to $n$. What's more, computing this integral does not need the decomposed boxes to have certain structures, contrasting the requirement of the grid structure in \citet{hupkens2015faster}. Hence, we can decompose the integral region into boxes using any of the previously discussed decomposition methods in Section~\ref{sec:EHVI}. 
	It is worth mentioning that usually $ \int_{box(\bm{u},\bm{l})} P(\bm{y}\prec\bm{y}')  \mathrm{d}\bm{y}' \neq  \int_{box(\bm{u},\bm{l})} \mathcal{H}_I( \bm{y}', A) P(\bm{y} = \bm{y}')  \mathrm{d}\bm{y}'$. The two integrals are only equal when the integral region is the whole nondominated region $U$.  
	

	\section{Decomposing Integral Regions}    \label{sec:integral region}
	
	With the closed-form integral computation for each hyperrectangle, we now integrate it with different decomposition methods to propose algorithms, including 
	1) grid-based algorithm, 2) WFG-based algorithm, and 3) CLM-based algorithm to calculate EHVI exactly given the nondominated set.
	
	
	\subsection{Grid-based Decomposition}
\begin{figure}[t!]
		\centering
		\vspace{-0.1in}
		\includegraphics[width = 3.25 in]{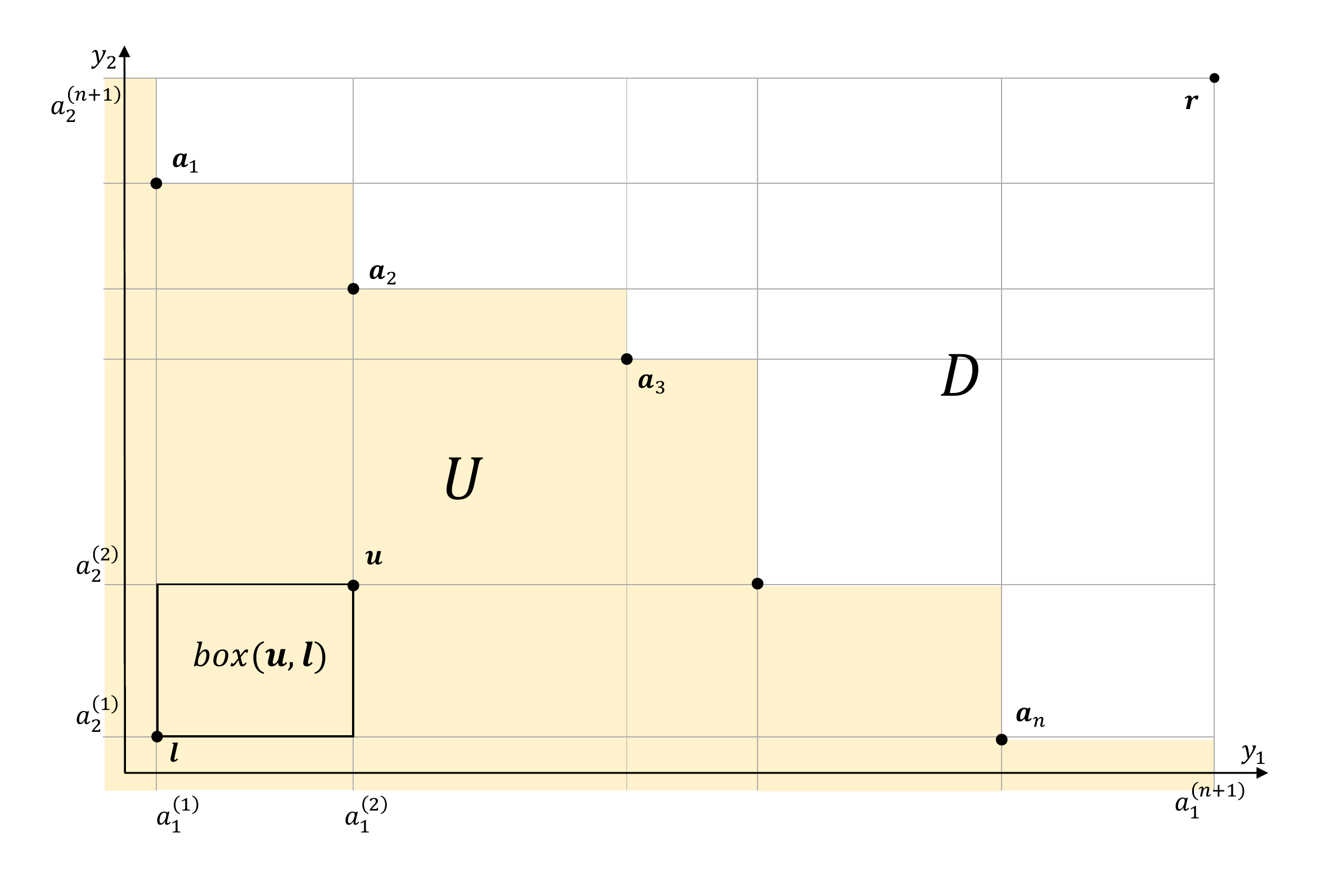}
		\vspace{-0.175in}
		\caption{Illustration of grid-based decomposition method: $U$ (yellow region) denotes the nondominated region; $D$ denotes the dominated region. The grid coordinates are indicated by the sorted lists $a_1^{(i)}$ and $a_2^{(i)}$ for the first and second objectives. }
		\label{fig:grid}
		\vspace{-0.15in}
	\end{figure}

	The grid-based algorithm decomposes the integral region into $O(n^m)$ $m$-dimensional disjoint boxes based on a grid structure, suggested by \citet{hupkens2015faster}. Denote the sorted list of $j^{th}$ coordinates of elements in nondominated set $A$ as $a_j^{(1)}, a_j^{(2)}, \ldots, a_j^{(n)}$, and assume $a_j^{(0)} = -\infty$, $a_j^{(n+1)} = r_j$. 
	We can then build the grid based on the sorted lists of $m$ objectives and generate boxes (cells) bounded by the grid as shown in Fig. \ref{fig:grid}. 
	
	
	The nondominated region can be decomposed into the generated  boxes with lower bound nondominated or belonging to $A$. Here is the proof.  
	Given a $box(\bm{u},\bm{l})$ and $A$, if the $\bm{l}$ is not dominated or equal to any vectors in $A$, i.e. $\bm{l} \nsucceq \bm{a}, \forall \bm{a} \in A$, then $\exists j$, s.t. $l_j < a_j$ and since $l_j$, $u_j$ and $a_j$ are from the same discrete sorting list, we have $u_j \le a_j$. 
	As a result the whole box bounded by $\bm{l}$ and $\bm{u}$ is nondominated by $A$. On the other hand, if $\bm{l}$ is dominated by one vector $\bm{a}\in A$, then the whole box is dominated by $\bm{a}$. Therefore the union of boxes with lower bound nondominated or belonging to $A$ equals to the nondominated set. 
	
	
	The relationship between lower bounds $\bm{l}$ and $A$ can be identified in constant time, by comparing $l_m$ with $H(l_1, l_2, \ldots, l_{m-1})$, an array indexed by the first $m-1$ coordinates which stores the lowest value of the   $m$-th coordinate given $\bm{l}\succeq A$. Array $H$ stores values of all possible combinations of the first $m-1$ coordinates of lower bounds, and the number of combinations, also the size of array, is $n^{m-1}$. The array can be built by going through all vectors $\bm{a}\in A$ in the descending order of the $m$-th coordinate, and for each $\bm{a}$, setting or updating $H(l_1, l_2, \ldots, l_{m-1}) = a_m$ for all $(l_1, l_2, \ldots, l_{m-1}) \succeq (a_1, a_2, \ldots, a_{m-1} )$. The complexity of building this array is $O(n^m)$. After that the complexity of identifying nondominated boxes and calculating the corresponding integrals is $O(m\cdot n^m)$. Hence the total complexity of the grid-based algorithm is also $O(m\cdot n^m)$.  
	
	\subsection{WFG-based Decomposition}
	
	The grid-based algorithm is intuitive and easy to implement, but the number of boxes can be large when $m$ is high. This is undesirable since the algorithm running time depends mostly on the number of boxes. In fact, the grid-based decomposition can be inefficient since some boxes can be combined into a larger box. To decompose the integral region into as few boxes as possible, we can borrow the idea in the exact algorithms for calculating the dominated hypervolume based on decomposition~ \citep{while2006faster, while2012fast, russo2014quick, jaszkiewicz2018improved, lacour2017box}. Among them we choose to extend the Walking Fish Group (WFG) algorithm \citep{while2012fast} to calculate EHVI since it's one of the fastest algorithms for calculating hypervolume \citep{jaszkiewicz2018improved}. 
	
\begin{figure}[t!]
		\vspace{-0.13in}
		\centering
		\includegraphics[width = 3.25 in]{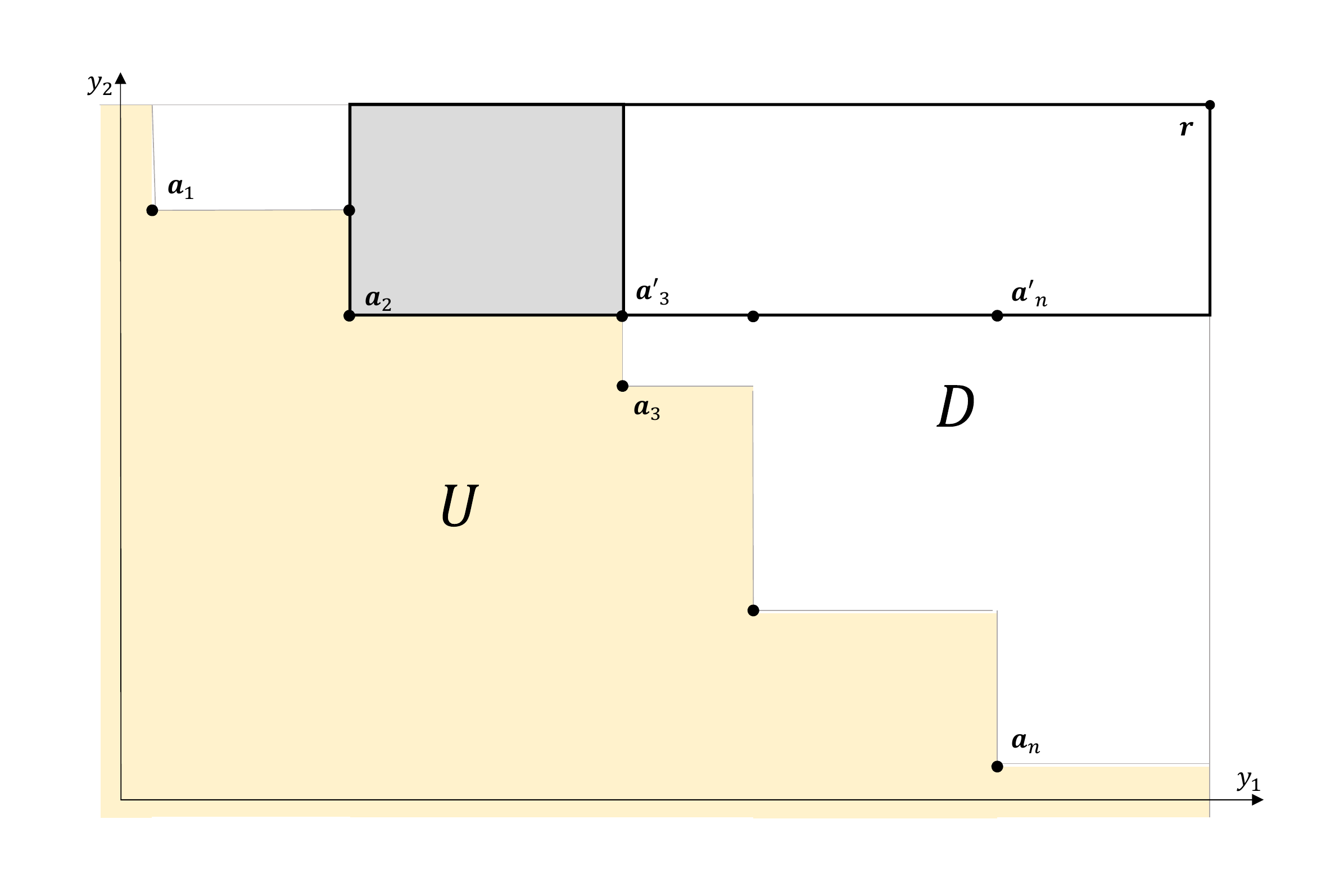}
		\vspace{-0.25in}
		
		\caption{One step in the WFG-based algorithm to calculate $\mathcal{H}_I(\bm{a}_2,\{\bm{a}_{3}, \ldots, \bm{a}_n\}) = \mathcal{H}(\{\bm{a}_2\}) - \mathcal{H}(\{\bm{a}'_{3}, \ldots, \bm{a}'_n\})$, denoted in the shaded region. }
		\label{figure:WFG}
		\vspace{-0.1in}
	\end{figure}

	WFG calculates the dominated hypervolume of a nondominated set $A$ as a summation of hypervolume improvements~(exclusive hypervolume):
	\begin{equation}
	\mathcal{H}(A) =\sum_{i = 1}^n  \mathcal{H}_I(\bm{a}_i, \{\bm{a}_{i+1}, \ldots, \bm{a}_n\}).
	\label{eq:exchypervolume}
	\end{equation}
	$\mathcal{H}_I(\cdot, \cdot)$ is defined in~\eqref{eq:HVI} and we define $\mathcal{H}_I(\bm{a}, \emptyset) = \mathcal{H}(\bm{a})$. 
	In~(\ref{eq:exchypervolume}), if we calculate the hypervolume improvements by the definition, we still need to calculate $\mathcal{H}(A)$ since $\mathcal{H}_I(\bm{a}_1, \{\bm{a}_2, \ldots, \bm{a}_n\}) = \mathcal{H}(A) - \mathcal{H}(\{\bm{a}_2, \ldots, \bm{a}_n\})$. Fortunately, the hypervolume improvement for a vector $\bm{a}$ with respect to $S$ is equal to the hypervolume of $\bm{a}$ minus the hypervolume of another set $S'$: 
	\begin{equation}
	\mathcal{H}_I(\bm{a}, S) = \mathcal{H}(\{\bm{a}\}) - \mathcal{H}(S'), 
	\label{eq:HVI3}
	\end{equation}		 
	where 
	$S' = \{\text{limit}(\bm{s}, \bm{a})|\bm{s}\in S\}$ and
	limit$(\bm{s}, \bm{a}) = (\max(s_1, a_1), \ldots, \max(s_m, a_m))$.
	The first term $\mathcal{H}(\{\bm{a}\})$ is just the hypervolume of $box(\bm{a}, \bm{r})$ and the second term $\mathcal{H}(S')$ can be calculated recursively by~\eqref{eq:exchypervolume}. By this recursive procedure, WFG transforms the dominated hypervolume into summation and difference of a series of boxes. Figure~\ref{figure:WFG} illustrates the calculation of $\mathcal{H}_I(\bm{a}_2, S = \{\bm{a}_{3}, \ldots, \bm{a}_n\}) = \mathcal{H}(\{\bm{a}_2\}) - \mathcal{H}(S' = \{\bm{a}'_{3}, \ldots, \bm{a}'_n\})$: the exclusive hypervolume of $\bm{a}_2$ is illustrated as the grey region. Note that, $\bm{a}'_3$ dominates all other vectors in $S'$ so that other vectors make no contribution to the hypervolume and can be disregarded. Before each recursion of calculating $\mathcal{H}(S')$, we can  take only the nondominated set of $S'$ into the following calculation, which makes WFG more efficient. It's shown in \citet{while2012fast} that in most cases the nondominated set will lose over 50\% vectors after being limited by one of its vectors. 
	It's worth mentioning that in high dimension cases, the exclusive hypervolumes are not always boxes, so the hypervolume improvement has to be calculated by~\eqref{eq:HVI3}. 
	
	To apply WFG to calculate EHVI, we can extend WFG to calculate the integral of integrand~\eqref{eq:8} over the dominated region $D$ and then subtract it from the integral of the whole region bounded by $\bm{r}$, which can be expressed by $\delta(box(-\bm{\infty}, \bm{r}))$ and calculated by \eqref{eq:boxclose}. 
	The dominated region can be decomposed into a series of boxes by following the same adding and subtracting rules in WFG for each box, for which we can compute the integral~\eqref{eq:EHVIbox} for each box. The complexity of the WFG-based algorithm is proportional to the number of decomposed boxes. In the worst case, the number of boxes can be $2^n - 1$, so the WFG-based algorithm has a worst case complexity $O(m\cdot 2^n)$. However, in most real-world cases the average complexity is much smaller, which will be demonstrated in Section~\ref{sec:experiment}. 
	
	\subsection{CLM-based Decomposition}
	
	
	We now derive another decomposition algorithm specially designed for three-objective EHVI calculation based on the CLM algorithm~\citep{beume2009complexity}. The CLM algorithm is a space sweeping algorithm along the $y_3$ axis, which achieves asymptotically optimal complexity $\Theta(n \log{n})$. It decomposes the 3d volume into slices with the volume of each slice being calculated by multiplying the height along the $y_3$ axis by the 2d area on the $(y_1, y_2)$ plane. The algorithm complexity is enhanced by taking advantage of a balanced binary search tree to store the nondominated projection on each plane. Such a tree structure can avoid repeated dominance checks and the recalculation of dominated area on the $(y_1, y_2)$ plane.
	
	We can use this algorithm to calculate the integral of integrand~\eqref{eq:8} in the dominated region $D$ and subtract it from the integral of the whole region to obtain EHVI, just as in our WFG-based algorithm. To calculate the integral, we only need to replace the volume calculation in the space sweeping algorithm with~\eqref{eq:EHVIbox}. Since~\eqref{eq:EHVIbox} also can be calculated in constant time, the CLM-based EHVI calculation will have the same asymptotically optimal algorithm complexity $\Theta(n\log{n})$ as in~\citet{beume2009complexity}. The existing KMAC algorithm~\citep{yang2017computing} uses a similar decomposition method extended from the CLM algorithm, also with the asymptotically optimal complexity. The performance comparison between KMAC and our CLM-based algorithm is found in the next section. 
	
	
	
	\section{Experiments}\label{sec:experiment}
	Experiments are performed to compare different algorithms calculating exact EHVI, including IRS \citep{hupkens2015faster}, CDD13 \citep{couckuyt2014fast}, KMAC \citep{yang2017computing} and our three algorithms with grid-based, WFG-based, and AVL-based decomposition algorithms.\footnote{IRS and KMAC are coded in C++ and obtained from \url{http://liacs.leidenuniv.nl/~csmoda/index.php?page=code}. We implemented the other algorithms in C/C++.} Based on the decomposition method, we can pair these algorithms as IRS and grid-based algorithm, CDD13 and WFG-based algorithm, KMAC and CLM-based algorithm. In the experiments, all the objectives are maximized, and the nondominated sets for EHVI calculation are based on the benchmark from \url{http://www.wfg.csse.uwa.edu.au/hypervolume/}. Each nondominated set $A$ is randomly generated, initially $A$ is empty, and sequentially a vector $\bm{a}$ randomly samples from box $[0.1, 10]^m$, $\bm{a}$ will be added to $A$ if $A\cup \bm{a}$ is still a nondominated set. The distribution of candidate points is assumed to be independent Gaussian with mean $\mu_i = 10$ and variance $\sigma_i = 2.5$, for $i \le m$. 
	
\begin{figure}[t!]
		\vspace{-0.05in}
		\centering
		\includegraphics[width = 3.25in]{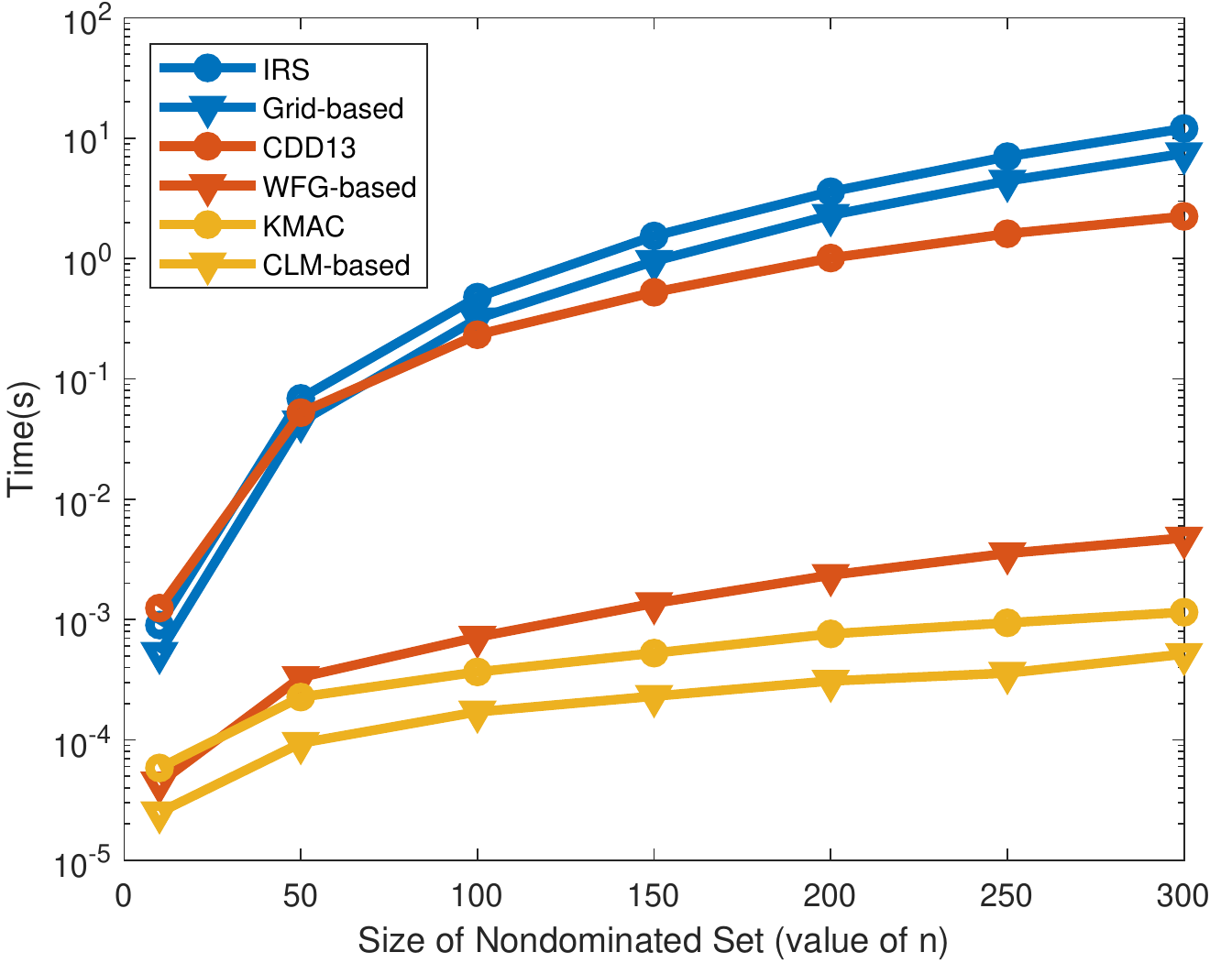}
		\vspace{-0.13in}
		\caption{Performance comparison for BO with respect to different $n$ for three-objective benchmark data ($m=3$). Reported computation time is averaged over five runs of 10 different benchmark nondominated sets.}
		\label{fig:3d}\vspace{-0.175in}
	\end{figure}

	The first experiment compares the performance of the proposed grid-based, WFG-based, and CLM-based algorithms with IRS, CDD13, and KMAC on the benchmark datasets with the nondominated set of size $n\in\{10, 50, 100, 150, 200, 250, 300\}$ when $m = 3$. For each data size $n$, the algorithms are applied on 10 different random nondominated sets, and perform 5 runs for each nondominated set. All algorithms were validated by producing the same
	results for the same nondominated sets. The average running time over different data size is shown in Fig. \ref{fig:3d}. The existing algorithms are marked with circles, and our algorithms with triangles. The algorithms with the same decomposition method are in the same color. From the figure, we can see that all of our proposed algorithms consistently perform better than the corresponding state-of-the-art algorithms with the same decomposition method. Among them, IRS and grid-based algorithms only have a constant difference in running time (ours is at least twice faster) because they have the same $O(n^3)$ complexity but IRS has a larger constant factor since it needs to calculate eight integral contribution terms for integral in each box compared with our proposed algorithm calculate a closed form expression. A similar trend can be observed between KMAC and our CLM-based algorithm, both have the same $\Theta(n \log n)$ complexity. Our WFG-based algorithm has a smaller overall complexity than CDD13 because for CDD13  calculating integral in each box requires summation over all boxes with time complexity higher than constant. Our WFG-based algorithm performs better than the grid-based algorithm, demonstrating that our WFG algorithm decomposes the integral regions into  fewer  boxes than the grid-based decomposition does. 

\begin{figure}[t!]
		\centering	\includegraphics[width = 3.25 in]{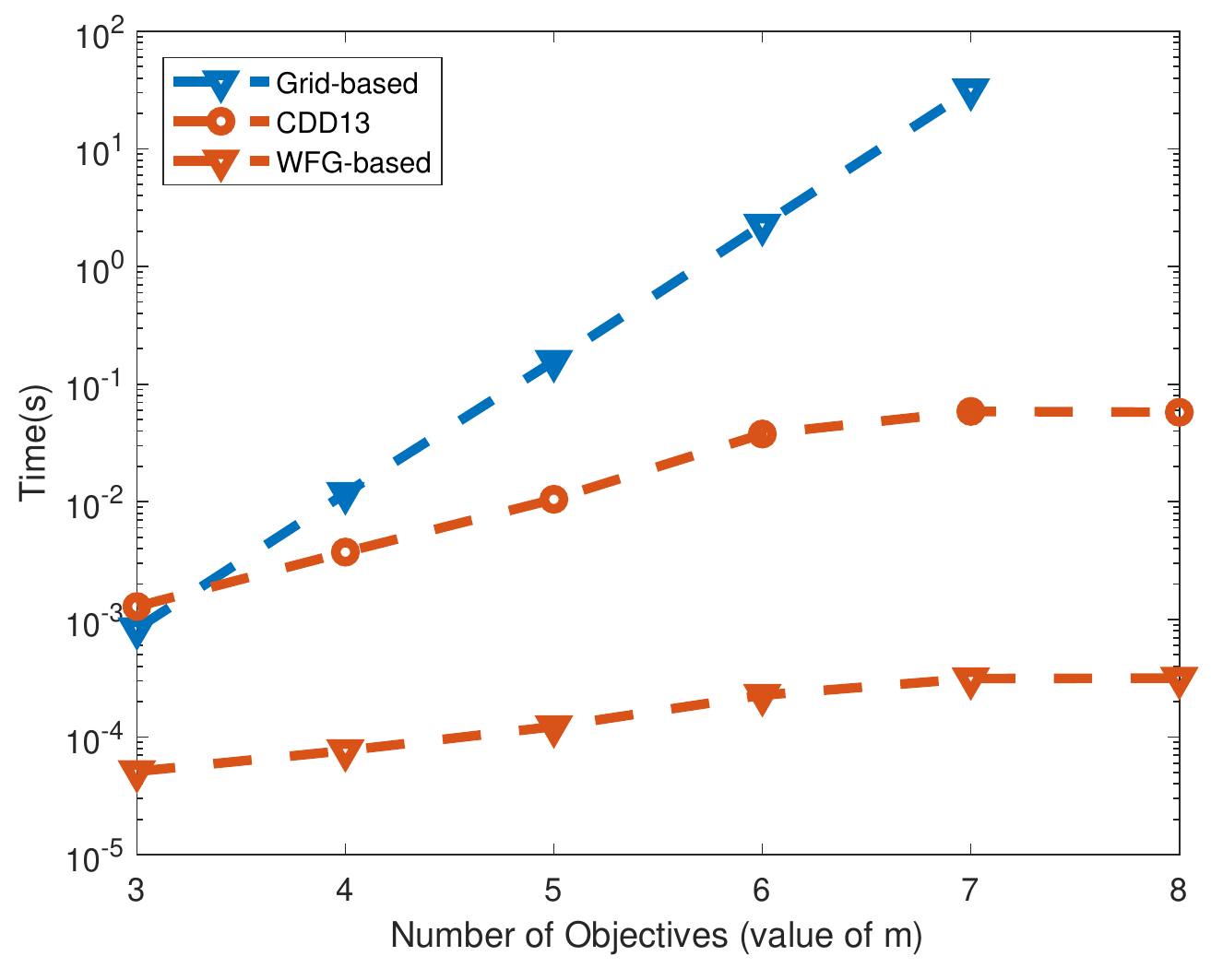}
		\vspace{-0.13in}
		\caption{Performance comparison with different $m$ between CCD13 and our grid-based and WFG-based algorithms based on the benchmark datasets of $n=10$. Reported computation time is averaged over five runs of 10 different benchmark nondominated sets.}
		\label{fig:10p} \vspace{-0.2in}
	\end{figure}

	In the second set of experiments, we show how the computation time varies with $m$. Since IRS, KMAC and our CLM-based algorithms are only applicable with $m=3$, we only compare our grid-based and WFG-based algorithms with CCD13 with the benchmark examples when $n = 10$ and  $m\in\{3, 4, 5, 6, 7, 8\}$. Again for each case we perform 5 runs of each algorithm on 10 different random nondominated sets and all the algorithm give same results for the same nondominated sets.  
	Figure~\ref{fig:10p} illustrates the increasing trend of the computation time with increasing $m$. When $m=8$, the grid-based algorithm takes much longer time ($>$ 500 seconds) and is not included in the plot for better visualization. From the figure, we can see the complexity of the grid-based algorithm is nearly exponential to the number of objectives, confirming the derived complexity $O(m\cdot n^m)$. The running time of our WFG-based algorithm and CDD13 is also nearly exponential to $m$ with a smaller bases~($<n$) when $m$ is small. But when $m$ approaches $n = 10$, the running time starts to saturate because the number of decomposed boxes by WFG algorithm is bounded by the worst case complexity $O(2^n)$.
	
	Last but not least, we have applied our CLM-based algorithm for implementing Bayesian experimental design to search for desirable NiTi Shape Memory Alloys (SMAs)~\citep{cox2017predictive, yu2015micromechanical}. SMAs are capable of undergoing reversible shape chages that make them ideal actuators. However, their properties are extremely sensitive to minute changes in their microstructure and composition and this opens up a very large design space that is extremely costly to explore solely through traditional experimental means.
	
	For our problem, we would like to search for NiTi SMAs in the materials design space $\mathcal{X}$ determined by the Ni concentration and precipitate Volume Fraction (VF). A Finite Element-based micromechanical model~\citep{cox2017predictive} is used to evaluate the material properties, from which we choose three objectives to optimize:
	1) the austenitic finish temperature being close to 30\textdegree{}C, i.e. $\min|A_f-30|$, 2) the difference of austenitic finish temperature and martensitic start temperature being close to 40\textdegree{}C, i.e.
	$\min |A_f-M_s-40|$, and 3) maximizing transformation strain. Here the martensitic temperatures are the ones that correspond to the reversible shape change upon cooling and the austenitic temperatures correspond to the reverse transformation upon heating.
	
	Independent Gaussian Processes with squared exponential kernel functions are adopted to model these objective functions. The hyperparameters of the Gaussian Processes are optimized by the quasi-Newton method. The initial set is generated by random sampling. Subsequently in each BO iteration, an exhaustive search over all unexplored points is performed to find the next query point optimizing EHVI.  With IRS, such an exhaustive search can cost more than one hour in each iteration, which is comparable to the running time of the micromechanical model. The expensive time cost for BO itself makes this impractical. On the other hand, with our new CLM-based algorithm the exhaustive search only takes a fraction of a second, orders of magnitude faster than the actual micromechanical simulations,  reducing significantly the cost associated to the optimal exploration of expensive design spaces. 
	
\begin{figure}[h!]
		\vspace{-0.01in}
		\centering
		\includegraphics[width = 3.25 in]{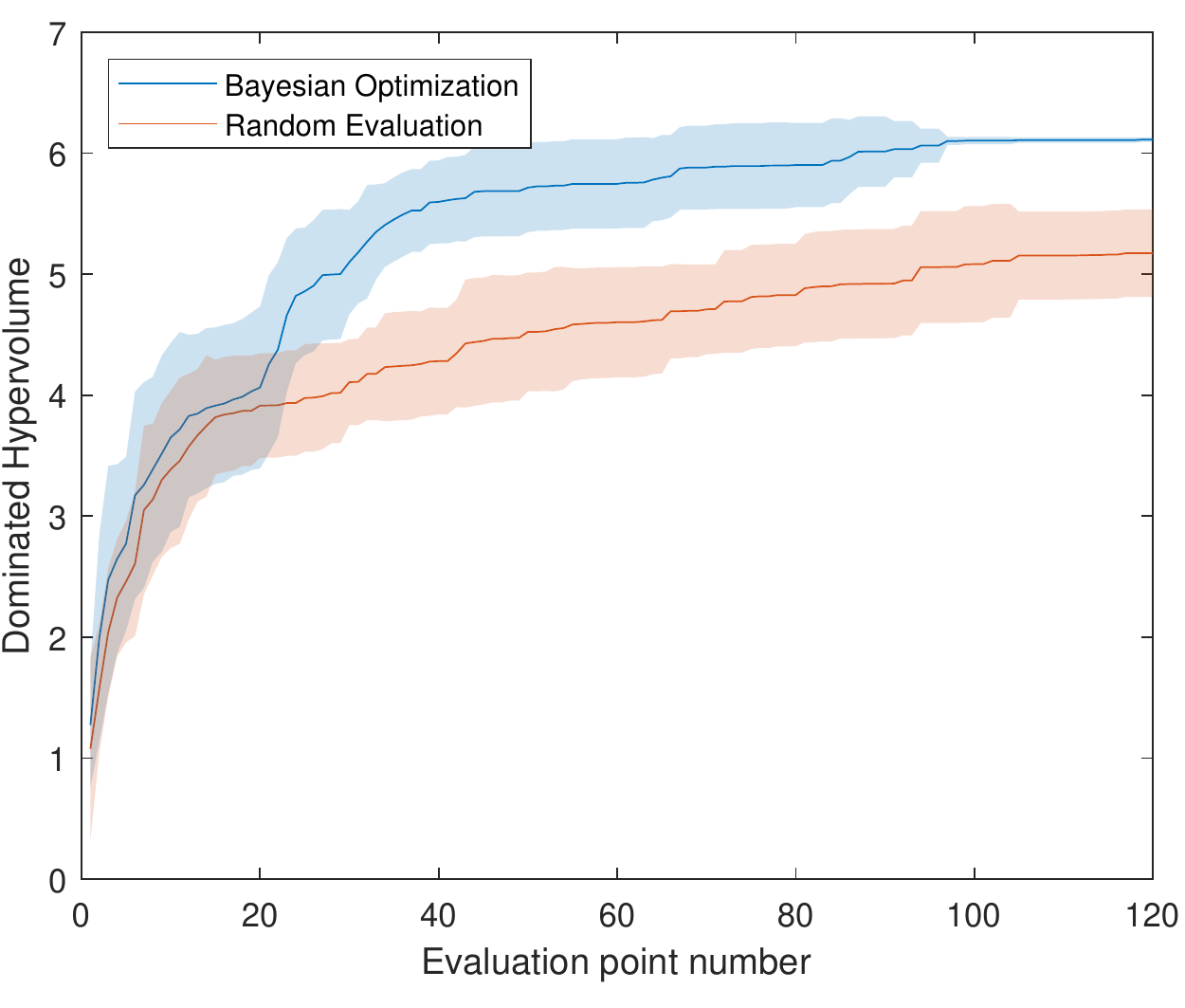}
		\vspace{-0.15in}
		\caption{Dominated volume change (average shown as the lines and standard deviation as shaded regions) over the iterations by Bayesian Optimization and Random Sampling. Each Bayesian optimization starts with an initial set of 20 random samples.}
		\label{fig:hypervolume}\vspace{-0.15in}
	\end{figure}

	We carried out 10 BO runs with the initial set of 20 samples and then sequential Bayesian experimental design with 100 following BO iterations. Performance of the sampling runs from Bayesian experimental design is assessed by the dominated hypervolume and compared against 10 runs with 120 random sampling evaluations. The performance comparison is shown in Fig. \ref{fig:hypervolume}. From the figure, we can see in the first 20 evaluations, there are barely no difference between the two sampling methods, since both methods are actually getting random samples. But after 20 initial evaluations, Bayesian experimental design performs much better than random sampling. After 100 evaluations, most Bayesian optimization runs begin to saturate with the variance converging to zero, demonstrating that the procedure has identified the Pareto front.   
	
	\section{Conclusions}
	
	We focus on improving the efficiency of calculating the exact EHVI, especially for cases with many objectives~($m\geq 3$). We have developed three fast exact EHVI calculation algorithms by integrating our new closed-form integral computation for hyperrectangles with three hypervolume decomposition methods. Experiments have shown that with the same decomposition method, our new algorithms are faster than state-of-the-art counterparts, also confirming the derived computation complexity. Moreover, our WFG-based algorithm can speed up the existing algorithm by more than two orders of magnitude when $m\geq 3$, enabling much more efficient Bayesian experimental design. In materials science, the need to carry out optimal sequential experiments over multi-objective spaces is necessary and approaches such as the ones demonstrated here empower autonomous materials discovery as demonstrated in our Shape Memory Alloy case study. This contribution, however, has a much wider applicability, beyond materials science.

	
	
	
	\section*{Acknowledgements}
	
	We would like to acknowledge the support from NSF through Grants No. 1534534 (\emph{DMREF: Accelerating the Development of High Temperature Shape Memory Alloys}), No. 1835690 (\emph{Elements: Software: Autonomous, Robust, and Optimal In-Silico Experimental Design Platform for Accelerating Innovations in Materials Discovery}) and No. 1553281(\emph{CAREER: Knowledge-driven Analytics, Model
		Uncertainty, and Experiment Design}). The authors also acknowledge Alexandros Solomou for providing the simulation code and data corresponding to the SMA BO demonstration. SMA micromechanical simulations were carried out at the Supercomputing Facility at Texas A\&M University. 

	\bibliography{algorithm}

\begin{thebibliography}{25}
\providecommand{\natexlab}[1]{#1}
\providecommand{\url}[1]{\texttt{#1}}
\expandafter\ifx\csname urlstyle\endcsname\relax
  \providecommand{\doi}[1]{doi: #1}\else
  \providecommand{\doi}{doi: \begingroup \urlstyle{rm}\Url}\fi

\bibitem[Beume et~al.(2009)Beume, Fonseca, L{\'o}pez-Ib{\'a}{\~n}ez, Paquete,
  and Vahrenhold]{beume2009complexity}
Nicola Beume, Carlos~M. Fonseca, Manuel L{\'o}pez-Ib{\'a}{\~n}ez, Lu{\'\i}s
  Paquete, and Jan Vahrenhold.
\newblock On the complexity of computing the hypervolume indicator.
\newblock \emph{IEEE Transactions on Evolutionary Computation}, 13\penalty0
  (5):\penalty0 1075--1082, 2009.

\bibitem[Cadman et~al.(2013)Cadman, Zhou, Chen, and Li]{cadman2013design}
Joseph~E. Cadman, Shiwei Zhou, Yuhang Chen, and Qing Li.
\newblock On design of multi-functional microstructural materials.
\newblock \emph{Journal of Materials Science}, 48\penalty0 (1):\penalty0
  51--66, 2013.

\bibitem[Chaloner and Verdinelli(1995)]{chaloner1995bayesian}
Kathryn Chaloner and Isabella Verdinelli.
\newblock Bayesian experimental design: A review.
\newblock \emph{Statistical Science}, pages 273--304, 1995.

\bibitem[Couckuyt et~al.(2014)Couckuyt, Deschrijver, and
  Dhaene]{couckuyt2014fast}
Ivo Couckuyt, Dirk Deschrijver, and Tom Dhaene.
\newblock Fast calculation of multiobjective probability of improvement and
  expected improvement criteria for pareto optimization.
\newblock \emph{Journal of Global Optimization}, 60\penalty0 (3):\penalty0
  575--594, 2014.

\bibitem[Cox et~al.(2017)Cox, Franco, Wang, Baxevanis, Karaman, and
  Lagoudas]{cox2017predictive}
A~Cox, B~Franco, S~Wang, T~Baxevanis, I~Karaman, and DC~Lagoudas.
\newblock Predictive modeling of the constitutive response of precipitation
  hardened ni-rich niti.
\newblock \emph{Shape Memory and Superelasticity}, 3\penalty0 (1):\penalty0
  9--23, 2017.

\bibitem[Emmerich and Fonseca(2011)]{emmerich2011computing}
Michael T.~M. Emmerich and Carlos~M. Fonseca.
\newblock Computing hypervolume contributions in low dimensions: asymptotically
  optimal algorithm and complexity results.
\newblock In \emph{International Conference on Evolutionary Multi-Criterion
  Optimization}, pages 121--135. Springer, 2011.

\bibitem[Emmerich and Klinkenberg(2008)]{emmerich2008computation}
Michael T.~M. Emmerich and Jan-willem Klinkenberg.
\newblock The computation of the expected improvement in dominated hypervolume
  of pareto front approximations.
\newblock \emph{Rapport technique, Leiden University}, 34, 2008.

\bibitem[Emmerich et~al.(2006)Emmerich, Giannakoglou, and
  Naujoks]{emmerich2006single}
Michael T.~M. Emmerich, Kyriakos~C. Giannakoglou, and Boris Naujoks.
\newblock Single-and multiobjective evolutionary optimization assisted by
  gaussian random field metamodels.
\newblock \emph{IEEE Transactions on Evolutionary Computation}, 10\penalty0
  (4):\penalty0 421--439, 2006.

\bibitem[Emmerich et~al.(2016)Emmerich, Yang, Deutz, Wang, and
  Fonseca]{emmerich2016multicriteria}
Michael T.~M. Emmerich, Kaifeng Yang, Andr{\'e} Deutz, Hao Wang, and Carlos~M.
  Fonseca.
\newblock A multicriteria generalization of bayesian global optimization.
\newblock In \emph{Advances in Stochastic and Deterministic Global
  Optimization}, pages 229--242. Springer, 2016.

\bibitem[Feliot et~al.(2017)Feliot, Bect, and Vazquez]{feliot2017bayesian}
Paul Feliot, Julien Bect, and Emmanuel Vazquez.
\newblock A bayesian approach to constrained single-and multi-objective
  optimization.
\newblock \emph{Journal of Global Optimization}, 67\penalty0 (1-2):\penalty0
  97--133, 2017.

\bibitem[Frazier et~al.(2009)Frazier, Powell, and
  Dayanik]{frazier2009knowledge}
Peter Frazier, Warren Powell, and Savas Dayanik.
\newblock The knowledge-gradient policy for correlated normal beliefs.
\newblock \emph{INFORMS journal on Computing}, 21\penalty0 (4):\penalty0
  599--613, 2009.

\bibitem[Hupkens et~al.(2015)Hupkens, Deutz, Yang, and
  Emmerich]{hupkens2015faster}
Iris Hupkens, Andr{\'e} Deutz, Kaifeng Yang, and Michael Emmerich.
\newblock Faster exact algorithms for computing expected hypervolume
  improvement.
\newblock In \emph{International Conference on Evolutionary Multi-Criterion
  Optimization}, pages 65--79. Springer, 2015.

\bibitem[Jaszkiewicz(2018)]{jaszkiewicz2018improved}
Andrzej Jaszkiewicz.
\newblock Improved quick hypervolume algorithm.
\newblock \emph{Computers \& Operations Research}, 90:\penalty0 72--83, 2018.

\bibitem[Lacour et~al.(2017)Lacour, Klamroth, and Fonseca]{lacour2017box}
Renaud Lacour, Kathrin Klamroth, and Carlos~M. Fonseca.
\newblock A box decomposition algorithm to compute the hypervolume indicator.
\newblock \emph{Computers \& Operations Research}, 79:\penalty0 347--360, 2017.

\bibitem[Luo et~al.(2015)Luo, Shimoyama, and Obayashi]{luo2015study}
Chang Luo, Koji Shimoyama, and Shigeru Obayashi.
\newblock A study on many-objective optimization using the
  kriging-surrogate-based evolutionary algorithm maximizing expected
  hypervolume improvement.
\newblock \emph{Mathematical Problems in Engineering}, 2015, 2015.

\bibitem[McAfee et~al.(2012)McAfee, Brynjolfsson, Davenport, Patil, and
  Barton]{mcafee2012big}
Andrew McAfee, Erik Brynjolfsson, Thomas~H. Davenport, DJ~Patil, and Dominic
  Barton.
\newblock Big data: the management revolution.
\newblock \emph{Harvard business review}, 90\penalty0 (10):\penalty0 60--68,
  2012.

\bibitem[Patel and Read(1996)]{patel1996handbook}
Jagdish~K. Patel and Campbell~B. Read.
\newblock \emph{Handbook of the normal distribution}, volume 150.
\newblock CRC Press, 1996.

\bibitem[Picheny(2015)]{picheny2015multiobjective}
Victor Picheny.
\newblock Multiobjective optimization using gaussian process emulators via
  stepwise uncertainty reduction.
\newblock \emph{Statistics and Computing}, 25\penalty0 (6):\penalty0
  1265--1280, 2015.

\bibitem[Provost and Fawcett(2013)]{provost2013data}
Foster Provost and Tom Fawcett.
\newblock Data science and its relationship to big data and data-driven
  decision making.
\newblock \emph{Big data}, 1\penalty0 (1):\penalty0 51--59, 2013.

\bibitem[Russo and Francisco(2014)]{russo2014quick}
Lu{\'\i}s M.~S. Russo and Alexandre~P. Francisco.
\newblock Quick hypervolume.
\newblock \emph{IEEE Transactions on Evolutionary Computation}, 18\penalty0
  (4):\penalty0 481--502, 2014.

\bibitem[While et~al.(2006)While, Hingston, Barone, and
  Huband]{while2006faster}
Lyndon While, Philip Hingston, Luigi Barone, and Simon Huband.
\newblock A faster algorithm for calculating hypervolume.
\newblock \emph{IEEE transactions on evolutionary computation}, 10\penalty0
  (1):\penalty0 29--38, 2006.

\bibitem[While et~al.(2012)While, Bradstreet, and Barone]{while2012fast}
Lyndon While, Lucas Bradstreet, and Luigi Barone.
\newblock A fast way of calculating exact hypervolumes.
\newblock \emph{IEEE Transactions on Evolutionary Computation}, 16\penalty0
  (1):\penalty0 86--95, 2012.

\bibitem[Wiering and Van~Otterlo(2012)]{wiering2012reinforcement}
Marco Wiering and Martijn Van~Otterlo.
\newblock Reinforcement learning.
\newblock \emph{Adaptation, learning, and optimization}, 12:\penalty0 51, 2012.

\bibitem[Yang et~al.(2017)Yang, Emmerich, Deutz, and
  Fonseca]{yang2017computing}
Kaifeng Yang, Michael T.~M. Emmerich, Andr{\'e} Deutz, and Carlos~M. Fonseca.
\newblock Computing 3-d expected hypervolume improvement and related integrals
  in asymptotically optimal time.
\newblock In \emph{International Conference on Evolutionary Multi-Criterion
  Optimization}, pages 685--700. Springer, 2017.

\bibitem[Yu et~al.(2015)Yu, Kang, and Kan]{yu2015micromechanical}
Chao Yu, Guozheng Kang, and Qianhua Kan.
\newblock A micromechanical constitutive model for anisotropic cyclic
  deformation of super-elastic niti shape memory alloy single crystals.
\newblock \emph{Journal of the Mechanics and Physics of Solids}, 82:\penalty0
  97--136, 2015.

\end{thebibliography}
	
\end{document}